# TENSOR OBJECT CLASSIFICATION VIA MULTILINEAR DISCRIMINANT ANALYSIS NETWORK

*Rui Zeng*[1, 4], *Jiasong Wu*[1, 2, 3, 4], *Lotfi Senhadji*[2, 3, 4], *Huazhong Shu*[1, 4]

[1]LIST, Key Laboratory of Computer Network and Information Integration (Southeast University), Ministry of Education, Nanjing 210096, China
[2]INSERM, U 1099, 35000 Rennes, France
[3]Laboratoire Traitement du Signal et de l'Image, Université de Rennes 1, Rennes, France
[4]Centre de Recherche en Information Biomédicale Sino-français (CRIBs)
E-mail: zengrui@seu.edu.cn, jswu@seu.edu.cn, lotfi.senhadji@univ-rennes1.fr, shu.list@seu.edu.cn

## ABSTRACT

This paper proposes a multilinear discriminant analysis network (MLDANet) for the recognition of multidimensional objects, known as tensor objects. The MLDANet is a variation of linear discriminant analysis network (LDANet) and principal component analysis network (PCANet), both of which are the recently proposed deep learning algorithms. The MLDANet consists of three parts: 1) The encoder learned by MLDA from tensor data. 2) Features maps obtained from decoder. 3) The use of binary hashing and histogram for feature pooling. A learning algorithm for MLDANet is described. Evaluations on UCF11 database indicate that the proposed MLDANet outperforms the PCANet, LDANet, MPCA + LDA, and MLDA in terms of classification for tensor objects.

## 1. INTRODUCTION

One key ingredient for the success of deep learning in visual content classification is the utilization of convolution architectures [1-3], which are inspired by the structure of human visual system [4]. A convolution neural network (CNN) [2] consists of multiple trainable stages stacked on the top of each other, following a supervised classifier. Each stage of CNN is organized in two layers: convolution layer and pooling layer.

Recently, Chan et al. [5] proposed a new convolutional architecture, namely, principal component analysis network (PCANet), which uses the most basic operation (PCA) to learn the dictionary in the convolution layer and the pooling layer is composed of the simplest binary hashing and histogram. The PCANet leads to some pleasant and thought-provoking surprises: such a basic network has achieved the state-of-the-art performance in many visual content datasets.

Meanwhile, Chan et al. [5] proposed linear discriminant analysis network (LDANet) as a variation of PCANet.

However, PCANet and LDANet are deteriorated when dealing with visual content, which is naturally represented as tensor objects. This is because when using PCANet or LDANet, the multidimensional patches, taken from visual content, are simply converted to vector to learn the dictionary. It is well known that vector representation of patches breaks the natural structure and correlation in the original visual content. Moreover, it may also, suffer from the so-called curse of dimensionality [6].

Recently, there is growing interest in the tensorial extension of deep learning algorithms. Yu et al. [7] proposed deep tensor neural network (DTNN), which can be seen as a tensorial extension of deep neural network (DNN). It was shown that DTNN outperforms DNN in large vocabulary speech recognition. Hutchinson et al. [8] presented the tensorial extension of deep stack neural network, which has been successfully used in MNIST handwriting image recognition, phone classification, etc. However, the similar tensorial extension research has not been reported for deep learning algorithms with convolutional architecture.

In this paper, we propose a simple deep learning algorithm for tensor object classification, that is, multilinear discriminant analysis network (MLDANet), which is a tensorial extension of PCANet and LDANet. The simulation on UCF11 database [9] demonstrates that the MLDANet outperforms PCANet and LDANet in terms of classification accuracy for tensor objects.

## 2. REVIEW OF MULTILINEAR DISCRIMINANT ANALYSIS

In this section, we briefly review MLDA [10], which is a multilinear extension of LDA. The MLDA obtains discriminative features through maximizing the fisher discrimination criterion, which is described as follows.

An $N$-th tensor object is denoted as $X \in \mathbb{R}^{I_1 \times I_2 \times \cdots \times I_N}$. It is addressed by $N$ indices $i_n$, $n = 1, 2, \ldots, N$, and each $i_n$ addresses the $n$-mode of $X$. The $n$-mode tensor product of $X$ by a matrix $U \in \mathbb{R}^{J_n \times I_n}$ is defined as:

$$(X \times_n U)_{(i_1,\ldots,i_{n-1},j_n,i_{n+1},\ldots,i_N)} = \sum_{i_n} X_{(i_1,\ldots,i_N)} \cdot U_{(j_n,i_n)}. \quad (1)$$

The projection from tensor $X \in \mathbb{R}^{I_1 \times I_2 \times \cdots \times I_N}$ to a scalar $y$ can be described as follows:

$$y = X \times_1 \mathbf{u}^{(1)T} \times_2 \mathbf{u}^{(2)T} \cdots \times_N \mathbf{u}^{(N)T}, \quad (2)$$

where $\mathbf{u}^{(n)T}, n=1,\ldots,N$ is a set of unit projection vectors. This tensor to scalar projection is called elementary multilinear projection (EMP), which consists of one projection vector in each mode. An EMP of a tensor $X \in \mathbb{R}^{I_1 \times I_2 \times I_3}$ is illustrated in Fig. 1.

The tensor-to-vector projection (TVP) from a tensor $X \in \mathbb{R}^{I_1 \times I_2 \times \cdots \times I_N}$ to a vector $\mathbf{y} \in \mathbb{R}^P$ is to find a vector set $\{\mathbf{u}_p^{(n)T}, n=1,\ldots,N\}_{p=1}^P$, which are able to do $P$ times EMP. The process can be described as:

$$\mathbf{y} = X \times_{n=1}^N \{\mathbf{u}_p^{(n)T}, n=1,\ldots,N\}_{p=1}^P, \quad (3)$$

whose $p$th component is obtained from the $p$th EMP as $\mathbf{y}_{(p)} = X \times_1 \mathbf{u}_p^{(1)T} \times_2 \mathbf{u}_p^{(2)T} \cdots \times_N \mathbf{u}_p^{(N)T}$. Fig. 2 shows the schematic plot for TVP.

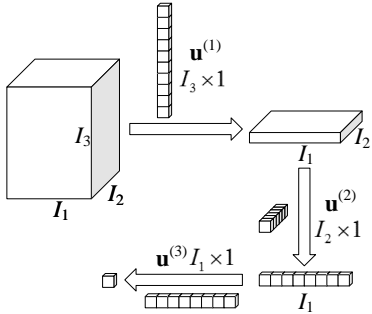

**Fig. 1**. The process of elementary multilinear projection (EMP).

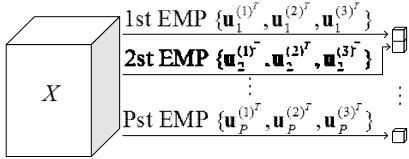

**Fig. 2**. Tensor-to-vector projection (TVP).

Suppose that we are given $M$ input tensor objects $\{X_m\}_{m=1}^M \in \mathbb{R}^{I_1 \times \cdots \times I_N}$, which contains $C$ classes. The $p$th projected scalar of $\{X_m\}_{m=1}^M$ are defined as $\{\mathbf{y}_{m_p}\}_{m=1}^M$, where $\mathbf{y}_{m_p} = X_m \times_{n=1}^N \{\mathbf{u}_p^{(n)T}\}_{n=1}^N$. So the between-class scatter matrix and the within-class scatter matrix for $p$th scalar tensor objects are defined as follows, respectively:

$$S_{B_p}^{\mathbf{y}} = \sum_{c=1}^C N_c (\bar{y}_{c_p} - \bar{y}_p)^2, \quad S_{W_p}^{\mathbf{y}} = \sum_{m=1}^M (y_{m_p} - \bar{y}_{c_{m_p}})^2. \quad (4)$$

where $\bar{y}_p = (1/M)\sum_m y_{m_p}$, $\bar{y}_{c_p} = (1/N_c)\sum_{m \in c} y_{m_p}$, $C$ is the number of classes, $N_c$ is the number of tensor objects in class $c$, $c_m$ is the class label for the $m$th tensor object. $\bar{y}_p = (1/M)\sum_m y_{m_p}$, and $\bar{y}_{c_p} = (1/N_c)\sum_{m \in c} y_{m_p}$. Thus, the Fisher's discriminant criterion for the $p$th scalar tensor objects is $F_p^{\mathbf{y}} = S_{B_p}^{\mathbf{y}} / S_{W_p}^{\mathbf{y}}$. The objective of MLDA is to determine a set of $P$ EMPs $\{\mathbf{u}_p^{(n)T}, n=1,\ldots,N\}_{p=1}^P$ satisfying the following conditions:

$$\{\mathbf{u}_p^{(n)T}, n=1,\ldots,N\} = \arg\max F_p^{\mathbf{y}} \quad (5)$$

## 3. THE ARCHITECTURE OF MLDANET

Fig. 3. Shows the architecture of MLDANet for third-order tensor objects classification. It contains two convolutional layers and one pooling layer. The filter bank in each convolutional layer is learned independently. We use binary hashing and histogram as pooling operation for the features extracted from the first two convolutional layers.

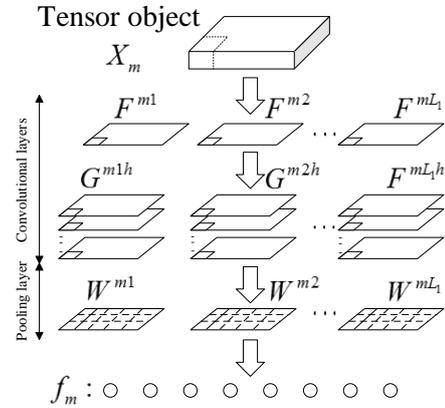

**Fig. 3**. The architecture of two-stage MLDANet.

### 3.1. The first stage of MLDANet

For the given $M$ third-order tensor objects $\{X_m\}_{m=1}^M \in \mathbb{R}^{I_1 \times I_2 \times I_3}$, which contains $C$ classes, we take tensor patches around the 1-mode and 2-mode by taking all 3-mode elements of $m$th tensor object, i.e., the tensor patch size is $k_1 \times k_2 \times I_3$, we collect all (overlapping) $I_1 \times I_2$ tensor patches from $X_m$. The tensor patches have the same class with $X_m$. We put these tensor patches into a set

$t_m = \{t_{m,q} \in \mathbb{R}^{k_1 \times k_2 \times I_3}\}_{q=1}^{I_1 \times I_2}$. By repeating the above process for every tensor objects, we can get all tensor patches $t = \{t_m\}_{m=1}^{M}$ for learning filter bank in the first stage.

Let $L_1$ be the number of filters in the first stage. We apply MLDA to $t$ to learn the $L_1$ vector sets $\{u_p^{(1)}, u_p^{(2)}, u_p^{(3)}\}_{p=1}^{L_1}$. For each tensor patch, we convert it into $L_1$ scalars by using $\{u_p^{(1)}, u_p^{(2)}, u_p^{(3)}\}_{p=1}^{L_1}$. Thus, the $l$th feature map of tensor object $X_m$ in the first stage is defined as:

$$F^{ml} = \mathrm{mat}(t_{m,q} \times_{n=1}^{3} \{u_l^{(n)T}\}) \in \mathbb{R}^{I_1 \times I_2}, q = 1, \ldots, I_1 \times I_2, \quad (6)$$

where $\mathrm{mat}(v)$ is a function that maps $v \in \mathbb{R}^{I_1 I_2}$ onto a matrix $F \in \mathbb{R}^{I_1 \times I_2}$.

For each tensor object, we can obtain $L_1$ feature maps of size $I_1 \times I_2$. We denote these feature maps of $m$th tensor object in the first stage with $\{F^{ml} \in \mathbb{R}^{I_1 \times I_2}, l = 1, \ldots, L_1\}_{m=1}^{M}$. The feature maps of each tensor object capture the main variation of original data.

### 3.2. The second stage of MLDANet

Through the first stage, the tensor object is already mapped into low-dimensional tensor feature, the dimensions of the 3-mode is much lower than that of the 1-mode and 2-mode. That is to say, the redundancy of 3-mode has been greatly reduced. Therefore, for the simplicity of computation and the convenience of building network, we use the conventional LDA in the second stage to learn the filter bank. The number of filters in the second stage is $L_2$.

Around each pixel, we take a $k_1 \times k_2$ patch, and collect all (overlapping) patches of all the feature maps $F^{ml}$, i.e., $\{r_{ml,q}\}_{q=1}^{I_1 \times I_2} \in \mathbb{R}^{k_1 \times k_2}$ where each $r_{ml,q}$ denotes the $q$th vectorized patch in the $l$th feature map of $m$th tensor object. We then subtract the patch mean from each patch, and construct the matrix for them $R_{ml} = [r_{ml,1}, r_{ml,2}, \ldots, r_{ml,I_1 \times I_2}]$, where $R_{ml}$ belongs to the same class with the $m$th tensor object. $S_c$ is the set of matrix $R_{ml}$ in class $c$. We then compute the class mean $\Gamma_c$ and the mean of class $\Gamma$ as follows:

$$\Gamma_c = \sum_{ml \in S_c} R_{ml} / |S_c|, \quad \Gamma = \sum_{\forall ml} R_{ml} / ML_1. \quad (7)$$

So, the within-class scatter matrix and the between-class scatter matrix are defined respectively as follows:

$$\begin{aligned} S_W &= \sum_{c=1}^{C} (\sum_{ml \in S_c} (R_{ml} - \Gamma_c)(R_{ml} - \Gamma_c)^T / |S_c|), \\ S_B &= \sum_{c=1}^{C} (\Gamma_c - \Gamma)(\Gamma_c - \Gamma)^T / C. \end{aligned} \quad (8)$$

We then get $w^* \in \mathbb{R}^{k_1 k_2 \times L_2}$ by maximizing the Fisher's discriminant criterion as follows:

$$w^* = \arg\max_w \frac{\mathrm{Tr}(w^T S_B w)}{\mathrm{Tr}(w^T S_W w)}, \text{ s.t. } w^T w = I_{L_2}.$$

By using mat function, each column of $w^*$ is converted into matrix $\{v_h \in \mathbb{R}^{k_1 \times k_2}\}_{h=1}^{L_2}$. These matrices are treated as the filter bank in the second stage. Let the $h$th output of the $l$th feature map for the $m$th tensor object in the second stage be

$$G^{mlh} = R_{ml} * v_h, h = 1, \ldots, L_2, l = 1, \ldots, L_1, \quad (9)$$

where $*$ denotes 2D convolution [2], and the boundary of $R_{ml}$ is zero-padded before convolving with $v_h$ so as to make $G^{mlh}$ have the same size as $R_{ml}$. The number of output feature maps of the second stage is $L_1 L_2$. One or more additional stages can be built if a deep architecture is found to be beneficial.

### 3.3. The pooling layer in MLDANet

First, we binarize each feature map by using Heaviside step function $H(\cdot)$, whose value are one for positive entries and zero otherwise. The binarized feature maps are denoted by $\bar{G}^{mlh} \in \{0,1\}^{I_1 \times I_2}$. Owing to every feature map capture different variations by $v_h$. $\bar{G}^{mlh}$ should be weighted to convert into a single integer-valued feature map:

$$W^{ml} = \sum_{h=1}^{L_2} 2^{h-1} \bar{G}^{mlh}. \quad (10)$$

Note that every entry of feature map $W^{ml}$ is integers in the range $[0, 2^{L_2}-1]$.

Next, we partition $W^{ml}$ into $B$ blocks, and then compute the histogram (with $2^{L_2}$ bins) of the decimal values in each block. All the $B$ histograms are concatenated into one vector as the $l$th feature vector $\mathrm{Bhist}(W^{ml})$ of tensor object $X_m$. The final feature of input tensor object $X_m$ is then defined as the set of feature vector, i.e.,

$$f_m = [\mathrm{Bhist}(W^{m1}), \mathrm{Bhist}(W^{m2}), \ldots, \mathrm{Bhist}(W^{mL_1})].$$

Note that the local block can be either overlapping or non-overlapping depending on applications [5].

## 4. EXPERIMENTAL RESULTS

We evaluate the performance of MLDANet on UCF11 dataset for tensor object classification. UCF11 is a sport action video dataset which contains 11 action categories: basketball shooting, biking, diving, golf swinging, horseback riding, soccer juggling, swinging, tennis swing, trampoline jumping, volleyball spiking, and walking with a dog. All videos in UCF11 are manually collected from YouTube and their sizes are all 240 × 320 pixels. For each category, the videos are grouped into 25 groups with more than 4 action clips in it. The video clips in the same group have common scenario. This video dataset is very challenging in classification due to large variations in camera motion, object appearance and pose, object scale, viewpoint, cluttered background, illumination, etc.

In this experiment, we only choose the first ten groups in each category. The total number of experimental videos is 642. For each group, half videos are randomly selected for training and others for testing. Every video is resized to 48 × 64 in order to reduce the computational complexity. Almost every video has variations in frames. For those frame larger than 20, we only choose the first twenty frames. For a few videos, whose frames are less than 20, we just copy the last frame to fill them.

We then compare the proposed MLDANet with PCANet [5], LDANet [5], MPCA+LDA [11], and MLDA [10]. The model parameters of MLDANet, PCANet, and LDANet all include the patch size $k_1 \times k_2$, the number of filters in each stage $L_1$, $L_2$, the number of stages, overlapping ratio of block, and the block size. Chan et al. [5] have shown that the appropriate number of filters $L_1$, $L_2$ in PCANet and LDANet is $L_1 = L_2 = 8$. By considering that MLDA is the tensorial extension of conventional LDA. Thus, we always set $L_1 = L_2 = 8$ for all networks. The patch size $k_1 \times k_2$ are changed from 3 × 3 to 7 × 7 and three box sizes 6 × 8, 12 × 16, 24 × 32 are considered here. The overlapping ratio is set to 50%. Unless stated otherwise, we use linear SVM classifier. The recognition rates of above networks averaged over 5 different random splits are shown in Figs. 4 (a)-(c).

the recognition accuracy of MPCA+LDA and MLDA in Fig. 4(d). The best performance of MLDANet, PCANet, LDANet, MPCA+LDA, and MLDA are listed in Table 1.

We see that all one-stage networks outperform two conventional tensor object classification algorithms, that is, MPCA + LDA and MLDA. The reason is that the convolutional architecture imitates the brain networks, which can provide more robust feature than other methods for visual content [4]. LDANet-1 achieves the best performance in the one-stage networks, but the improvement from LDANet-1 to LDANet-2 is not larger as that of MLDANet. PCANet-1 performs worse than those based on LDA algorithm networks like LDANet and MLDANet. It is because that LDA type algorithm makes the features which have the best classification performance, however, PCA maximize the directional variation in the features. MLDANet-1 is not as good as LDANet-1 because the feature extracted from MLDANet-1 is not appropriate as the direct input of linear SVM. For two-stage networks, MLDANet-2 achieves the best performance. Surprisingly, the performance of PCANet-2 increases not more than 18.24% compared to that of PCANet-1, but it is better than that of LDANet-2.

**Table 1**. The best performance of MLDANet, LDANet, PCANet, MPCA+LDA and MLDA.

| Methods | Accuracy |
|---|---|
| MLDANet-1 | 64.55 |
| MLDANet-2 | **78.93** |
| LDANet-1 | 73.58 |
| LDANet-2 | 76.59 |
| PCANet-1 | 58.68 |
| PCANet-2 | 76.92 |
| MPCA+LDA | 45.15 |
| MLDA | 38.46 |

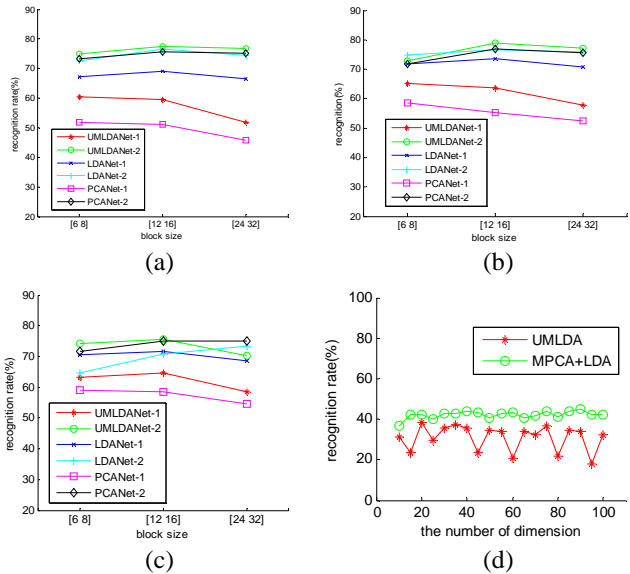

**Fig. 4**. Recognition rate overview in terms of the parameters of network chosen: (a) patch size 3 × 3. (b) patch size 5 × 5. (c) patch size 7 × 7 and (d) are the performance of MPCA+LDA and MLDA.

For conventional tensor object classification by using MPCA+LDA, we change the dimensions of input feature of LDA from 10 to 100. The dimensions of feature vector extracted from MLDA are changed from 10 to 100. We draw

## 5. CONCLUSIONS

In this paper, we have proposed and implemented a novel deep learning architecture, that is, MLDANet, which takes full advantage of the structure information in tensor objects by convolutional architecture. MLDANet is composed of two convolutional layers, which use MLDA and LDA to learn filter banks respectively and one pooling layer. We have evaluated the performance of the MLDANet on UCF11 and show that our model performs well in tensor object classification. This work provides the inspiration for other convolutional deep architectures in tensor object classification. As future works, we will focus on the tensorial extensions of CNN.

**ACKNOWLEDGEMENT**
This work was supported by the National Basic Research Pr ogram of China under Grant 2011CB707904, by the NSFC

under Grants 61201344, 61271312, 11301074, and by the SRF DP under Grants 20110092110023 and 20120092120036, the Project-ponsored by SRF for ROCS, SEM, and by Natural Science Foundation of Jiangsu Province under Grant BK2012329 and by Qing Lan Project. This work is supported by INSERM postdoctoral fellowship.